\def\year{2019}\relax
\documentclass[letterpaper]{article} 
\usepackage{aaai19}  
\usepackage{times}  
\usepackage{helvet}  
\usepackage{courier}  
\usepackage{url}  
\usepackage{graphicx}  
\usepackage{multirow}
\usepackage{amsfonts}
\usepackage{CJKutf8}
\usepackage{epsfig}
\usepackage{color}
\usepackage{graphicx}
\usepackage{amsfonts}
\usepackage{amsmath}
\usepackage{bm}
\begin{document} 
%
\title{Modeling Coherence for Discourse Neural Machine Translation}
\author{Hao Xiong, Zhongjun He, Hua Wu and Haifeng Wang\\
Baidu Inc. No. 10, Shangdi 10th Street, Beijing, 100085, China\\
{\{xionghao05, hezhongjun, wu\_hua, wanghaifeng\} }@baidu.com\\
}

\maketitle
\begin{CJK}{UTF8}{gbsn}
\begin{abstract}

Discourse coherence plays an important role in the translation of one text. However, the previous reported models most focus on improving performance over individual sentence while ignoring cross-sentence links and dependencies, which affects the coherence of the text. In this paper, we propose to use discourse context and reward to refine the translation quality from the discourse perspective. In particular, we generate the translation of individual sentences at first. 
Next, we deliberate the preliminary produced translations, and train the model
to learn the policy that produces discourse coherent text by a reward teacher. 
Practical results on multiple discourse test datasets indicate that our model significantly improves the translation quality over the state-of-the-art baseline system by +1.23 BLEU score. Moreover, our model generates more discourse coherent text and obtains +2.2 BLEU improvements when evaluated by discourse metrics.  

\end{abstract}
\section{Introduction}
Discourse coherence, such as relevant conjunction for adjacent sentences, plays an important role in the translation of the text. However, standard Neural Machine Translation (NMT) models \cite{sutskever2014sequence,bahdanau2014neural} mainly focus on improving translation quality over individual sentence, and ignoring cross-sentence links and dependencies. With this preference, the translation of each sentence is independent of the other sentences, thus there is no guarantee to generate discourse coherent text. 

Table \ref{tbl:inst} shows a concrete example from TED talks, where each sentence is rationally translated from the sentence perspective, but the missing conjunction `And', and the missing coreference for the predicate `build' cause the incoherence and poor readability of the entire text. 

Intuitively, to generate better translation of the entire text, the model 
deserves considering the cross-sentence connections and dependencies, generating discourse coherent translations. Towards this demand, most previous work \cite{wang2017exploiting,P18-1117} proposed to explore additional context, generally is certain preceding adjacent sentences, to reinforce the model. However, the major goal of these models is still the quality of individual sentence while not the entire text, thus it is hard for them to generate promising discourse coherent translations. 

 \begin{table}[t]

\begin{center}
\begin{tabular}{p{0.9 cm}|p{0.8\columnwidth}}

\textit{Source} & 我们加入霓虹 我们加入柔和的粉蜡色 我们使用新型材料。  \\

  &人们爱死这样的建筑了。 \\
   & 我们不断地建造。 \\
\hline
\textit{Ref} & We add neon  and we add pastels  and we use new materials.\\
& \underline{And} you love \underline{it}.\\
 & \underline{And} we can't give you enough of \underline{it}. \\
 \hline
 \textit{NMT} & We add the neon, we add soft, flexible crayons, and we use new materials. \\
  & [\textit{conj}]$_{miss}$People love architecture. \\
  & [\textit{conj}]$_{miss}$We keep building [\textit{coref}]$_{miss}$. \\
  
\end{tabular}
\end{center}
\caption{Instance of translation for one text consists of three sentences, where [\textit{conj}]$_{miss}$ indicates missing cross-sentence conjunction, and [\textit{coref}]$_{miss}$ indicates missing coreference. Although from the sentence perspective, the translations of second and third sentences produced by \textit{NMT} system are acceptable, however the fluency of the entire text is poor. }
\label{tbl:inst}
\end{table}

To address this problem, we take an insight into human translation behavior for one text, where we first translate each sentence independently, and then take some modifications towards making the entire text coherently and fluently, such as replacing conjunctions and keeping the translation of terminologies consistently. Ideally, this procedure can be divided into two processes: 1) generating preliminary translation of each sentence, 2) deliberating each translation with the satisfaction of discourse coherence. 

Motivated by the human translation behavior and the success of Deliberation Networks \cite{xia2017deliberation}, we propose a two-pass decoder translation model, aiming at improving the coherence of the entire text. 
Specifically, we generate the preliminary translation of each sentence using the canonical NMT model, and then employ the Deliberation Networks to refine the translations over the entire text. 
However, since the standard NMT models focus on fine-tuning on local $n$-gram patterns, training by maximum likelihood estimation, the produced translations are generally locally coherent. 

Intuitively, to generate discourse coherent translations, it requires training the model to learn the policy that receives more discourse coherent rewards straightforwardly.
To satisfy this requirement, we explore a novel measure to estimate the quality of discourse coherence, namely a reward teacher \cite{bosselut2018discourse}, which learns the ordering structure in one text trained by a bidirectional recurrent neural networks (biRNN) \cite{schuster1997bidirectional}. Rewarding by the reward teacher, the model learns to generate the discourse coherent translations while maintaining accurate translation for each individual sentence.

In particular, we design our model based on the Transformer architecture \cite{vaswani2017attention} according to its superior performance on machine translation tasks. 

We evaluate the performance of our model on the IWSLT speech translation task with TED talks \cite{cettoloEtAl:EAMT2012} as training corpus, which includes multiple entire talks. Practical experiments reveal that our model improves the sentence translation quality by +1.23 BLEU \cite{papineni2002bleu} score over one strong baseline. Moreover, when evaluated by discourse metrics, our model achieves an average improvements by +2.2 points in term of BLEU and +1.98 of METEOR \cite{denkowski:lavie:meteor-wmt:2014}. Through extensive experimental analysis, we confirm that our model can generate more discourse coherent translations than the baseline system.

To our knowledge, this is the first work on modeling coherence for discourse neural machine translation. The contributions of this paper can be concluded into the followings:
\begin{itemize}
\item We propose a two-pass decoder translation model to refine the translation of the discourse text.
\item We propose a policy learning technique that encourages the model generating coherent and fluent discourse translations.
\item We conduct extensive experiments to validate the effectiveness of our model, and analyze the results to confirm that our model can generate discourse coherent translations.
\end{itemize}

\section{Our Approach}
\label{sec:2}

Intuitively, it is plausible that a model with external context summarized from entire text, trained by a discourse-aware reward can generate more accurate and discourse coherent translations. Towards exploring global context to refine the translation, we take inspirations from the work of Deliberation Network, to translate sentence in one text independently by the first-pass decoder, and then to summarize the first-pass translation as the external context for the second-pass decoder. Although most existing related work summarizes the external context with gold sequence, here we follow the original method of Deliberation Network that takes the predicted sequence as our global context, since it can also potentially alleviate the \textit{exposure bias} problem \cite{ranzato2015sequence}.     

To let the NMT model be prone to generate discourse coherent text, we learn the recent advances in text generation task \cite{bosselut2018discourse}, and propose to use the overall ordering structure of a document as an approximation of discourse coherent. As the two-pass decoders learn the policy that tends to receive more discourse rewards from the reward teacher, it is possible for the model satisfying the above mentioned requirements for the discourse translation.

\subsection{Overall Architecture}
\label{sec:2.1}
In recent work, \citeauthor{vaswani2017attention}\shortcite{vaswani2017attention} proposed an effective encoder-decoder based translation model, namely Transformer. The Transformer follows an encoder-decoder architecture using stacked self-attention and fully connected layers for both the encoder and decoder.
In contrast to recurrent models, it avoids recurrence completely and drops the usage of complex Long Short-term Memory (LSTM) \cite{hochreiter1997long} and Gated Recurrent Unit (GRU) \cite{cho2014learning}, being more parallelizable and faster to train. According to its effectiveness and superior performance on the translation task, we develop our architecture based on the Transformer model.

\begin{figure*}[ht]
\centering
\includegraphics[width=6.5in]{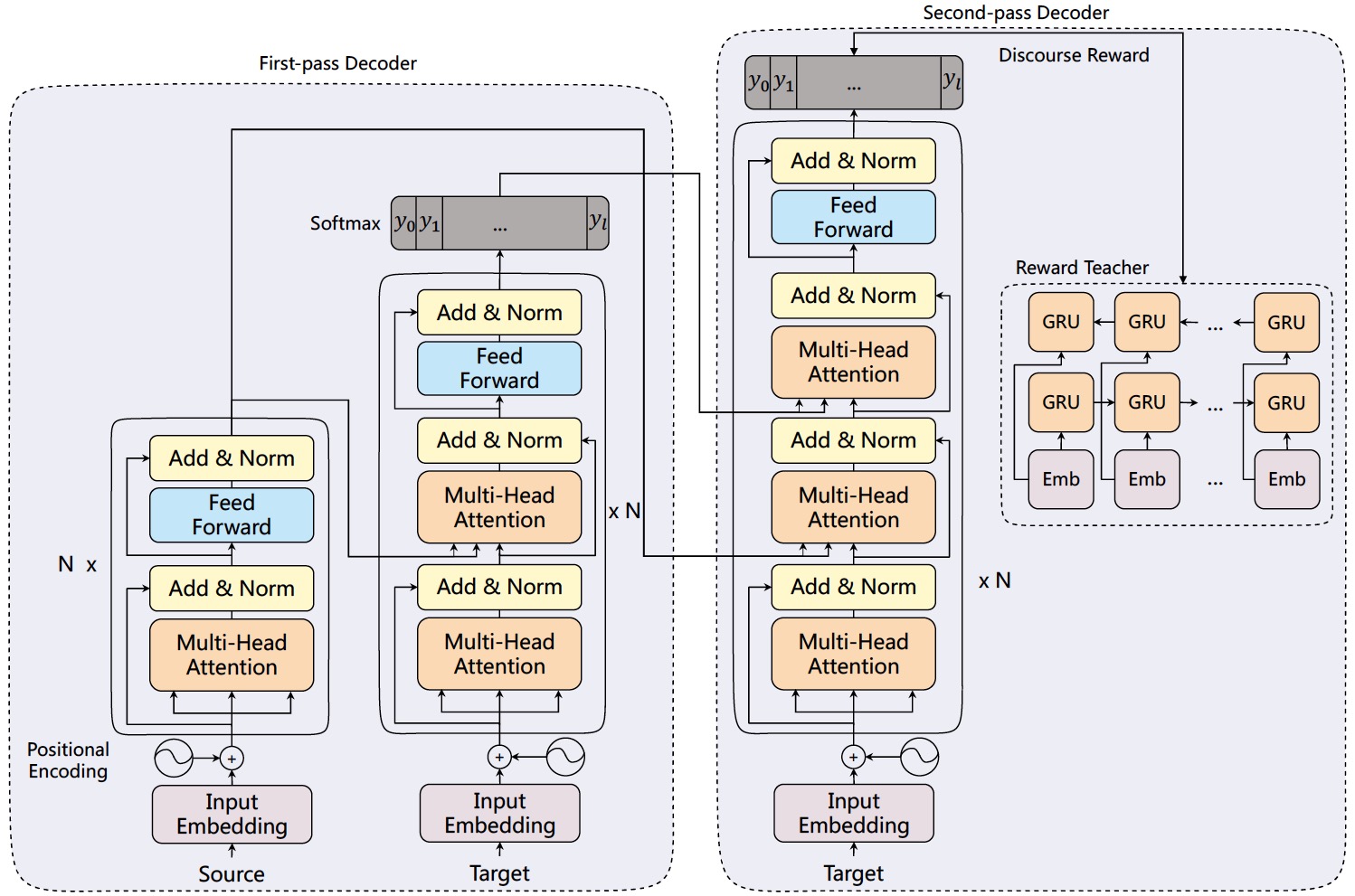}

\caption{Illustration of the overall architecture of our two-pass decoder translation model. The first-pass decoder produces translations as the canonical Transformer model does. And the second-pass decoder utilizes an additional self-attention layer, exploring more contextual information from the other sentences generated by the first-pass decoder. 
We also let the model learning the policy to generate more fluent and coherent translation by rewarding from a reward teacher. }
\label{fig:arch}
\end{figure*}

Figure \ref{fig:arch} illustrates the overall architecture of our proposed two-pass decoder translation model. 
It is clear that our model can be divided into three important components: the first-pass decoder, the second-pass decoder and the discourse reward teacher.    
Since the original architecture of the Transformer is well designed, fine-tuning on the quality of sentence translation, we replicate the original Transformer encoder-decoder architecture in the first-pass decoder. 

As mentioned in the literature \cite{P18-1117}, techniques developed for the recurrent based models did not present effective for the Transformer, instead they proposed two separated encoders to learn the representations for the context and source sequence respectively. Notably, the context used in their model is one preceding adjacent sentence appeared in the document. However, after preliminary experiments, we found that mechanically replication of previous techniques yields bad performance in our architecture. Instead we stack an additional self-attention layer in the canonical Transformer decoder, resulting in three types of self-attention layers in the second-pass decoder. Ideally, the additional self-attention layer will summarize the context from the translation of overall document generated by the first-pass decoder, making the decoder to generate discourse coherent translation as possible, since it has learned the potential translations of other sentences in this document.  

The last important component of our model is the discourse reward teacher, an offline trained biRNNs that rewards the model to generate more discourse coherent translation.
Before we describe the discourse reward teacher in details, we firstly draw some definitions of our two-pass decoder for better understanding of this paper.

\subsection{First-pass Decoder}
\label{sec:2.2}
We define a source document of $n$ sentences as $S_x = \{s_{x:0}, ..., s_{x:n-1}\}$ where each sentence $s_{x:i}$ have $T_i$ words. The goal of first-pass decoder is trained to minimize the negative log-likelihood of predicting the target word, $y_t$:
\begin{equation}
L_{mle1}= -\sum_{i}^{n}{\sum_{t}^{T_i}{\mathrm{log}P(y_t|y_0, ..., y_{t-1}, \mathrm{H}_{enc}, \mathrm{H}_{dec1})}}
\label{eq:mle1}
\end{equation}
where $T_i$ is the length of the target sequence, and the $\mathrm{H}_{enc}$, $\mathrm{H}_{dec1}$ are the representations of the encoder and the decoder in the first-pass decoder. 

\subsection{Second-pass Decoder}
\label{sec:2.3}
In contrast to the original Deliberation Network, where they proposed a complex joint learning framework to train the model. By virtue of our preliminary experiments, we found that a simpler solution is enough to obtain promising results. Thus, we treat the first-pass decoder and the second-pass decoder as two associative learning tasks by sharing the identical encoder, minimizing the following loss:
\begin{equation}
L_{mle}=L_{mle1}+L_{mle2}
\label{eq:mle}
\end{equation}
where 
\begin{equation}
L_{mle2}= -\sum_{i}^{n}{\sum_{t}^{T_i}{\mathrm{log}P(y_t|y_0, ..., y_{t-1}, \mathrm{Y}^{'}, \mathrm{H}_{enc}, \mathrm{H}_{dec2})}}
\label{eq:mle2}
\end{equation}
where $\mathrm{H}_{dec2}$ is the representation of the decoder in the second-pass decoder. Here we use different training parameters for decoders in the first-pass and the second-pass decoding stage independently. As in the first-pass decoder, we can use a greedy strategy or beam search to generate the preliminary document translations, $\mathrm{Y}^{'}$. Generally, $\mathrm{Y}^{'}$ is represented by word embeddings.   
\subsection{Reward Teacher}
\label{sec:2.4}
In recent work, \citeauthor{bosselut2018discourse}\shortcite{bosselut2018discourse} proposed two neural teachers, absolute order teacher and relative order teacher to model the coherence of generated text towards the text generation task. 
Motivated by their success, we extend this approach to the machine translation task, since in our task we need also generate coherent translations which is the same as in the text generation task. 

Specifically, we develop absolute order teacher as our reward teacher by training a sentence encoder to minimize the similarity between a sequence encoded in its forward order, and the same sequence encoded in the reverse order. 
\footnote{Although \citeauthor{bosselut2018discourse}\shortcite{bosselut2018discourse} obtains better results with the relative order teacher in their experiments, we found in the machine translation task an absolute order teacher is appropriate to obtain promising performance and omits the descriptions of the relative order teacher.}
Following the same definitions as \citeauthor{bosselut2018discourse}\shortcite{bosselut2018discourse}, 
each target sentence $s_{y:i}$ that has $L_i$ words, is represented as:
\begin{equation}
s_{y:i} = \sum_j^{L_i}{y_{ij}}
\label{eq:si}
\end{equation}
where ${y_{ij}}$ is a word embedding and $s_{y:i}$ is a sentence embedding. 

Identical to the canonical encoder in the recurrent models, each $s_{y:i}$ is passed to a GRU:
\begin{equation}
h_i=\mathrm{GRU}(h_{i-1}, s_{y:i})
\label{eq:gru}
\end{equation}
 and the final hidden state of the RNN is utilized as the representation of the document:
\begin{equation}
f(\mathrm{S_y})=h_n
\label{eq:gru}
\end{equation}
where $f(\mathrm{S_y})$ is the representation of the document and $h_n$ is the final state of the recurrent neural networks.

Intuitively, if one text is well organized, the similarity between the sentence embedding from reading the
sentences in the forward order, and from reading
the sentences in the reverse order, should be minimized. 

Thus, the absolute order teacher is trained to minimize the cosine similarity between two orders:
\begin{equation}
L_{abs}=\frac{\langle f(\overrightarrow{\mathrm{S_y}}), f(\overleftarrow{\mathrm{S_y}}) \rangle}{\Vert f(\overrightarrow{\mathrm{S_y}}) \Vert \Vert f(\overleftarrow{\mathrm{S_y}}) \Vert}
\label{eq:abs}
\end{equation}  
After training on the monolingual corpus, we use this learned teacher
to generate a reward that judges the generated sequence’s
ordering similarity to the gold sequence.

Notably, the reward teacher is
trained offline on gold sequences in an unsupervised
manner prior to training the NMT model, and its
parameters are fixed during policy learning.
 
\subsection{Policy Learning}
\label{sec:2.5}
Since training the two-pass decoder with maximum likelihood estimation produces translations
that are locally coherent, so there is no guarantee to generate coherent discourse. 
As mentioned in the above section, we train a reward teacher to reward the model that generates good ordering structure, encouraging the model to learn a policy that
produces discourse coherent translation explicitly. 
In this paper, we learn a policy using the
self-critical training approach \cite{rennie2017self}.

Specifically, for each training example (one document), a sequence $\hat{y}$ is generated by sampling from the models's distribution 
$P(\hat{y}_t|\hat{y}_0, ..., \hat{y}_{t-1}, \mathrm{Y}^{'}, \mathrm{H}_{enc}, \mathrm{H}_{dec2})$ of the second-pass decoder. Another sequence $y^*$ is generated by argmax decoding from 
$P(y^*_t|y^*_0, ..., y^*_{t-1}, \mathrm{Y}^{'}, \mathrm{H}_{enc}, \mathrm{H}_{dec2})$ at each time step $t$. 

The model is trained to minimize:
\begin{equation}
\begin{split}
L_{rl2} &= -\sum_{i}^{n}{\sum_{t}^{T_i}{R \cdot \mathrm{log}P(y_t|y_0, ..., y_{t-1}, \mathrm{Y}^{'}, \mathrm{H}_{enc}, \mathrm{H}_{dec2})}} \\
R &= r(\hat{y}) - r(y^*)
\end{split}
\label{eq:rl2}
\end{equation}
where $r(y^*)$ is the reward produced by the reward teacher
for the greedily decoded sequence, and $r(\hat{y})$ is the reward for the sampled sequence.

Since
$r(y^*)$ can be viewed as a baseline reward, the model learns to generate sequence that receives
more reward from the teacher than the best
sequence, which can be greedily decoded from the
current policy. This approach allows the model
to explore sequence that yields higher reward than
the current best policy.

Notably, here we introduce the policy learning for the second-pass decoder. In actual, it is natural facilitating the first-pass decoder with the identical technique, which will be described later.
\subsection{Absolute Order Reward}
Once the sequences $\hat{y}$, $y^*$ are generated, we use the absolute order reward teacher to reward these sequences:
\begin{equation}
\begin{split}
r(\hat{y}) &=\frac{\langle f(\overrightarrow{\mathrm{S_{\hat{y}}}}), f(\overrightarrow{\mathrm{S_y}}) \rangle}{\Vert f(\overrightarrow{\mathrm{S_{\hat{y}}}}) \Vert \Vert f(\overrightarrow{\mathrm{S_y}}) \Vert} - \frac{\langle f(\overrightarrow{\mathrm{S_{\hat{y}}}}), f(\overleftarrow{\mathrm{S_y}}) \rangle}{\Vert f(\overrightarrow{\mathrm{S_{\hat{y}}}}) \Vert \Vert f(\overleftarrow{\mathrm{S_y}}) \Vert}\\
r(y^*) &=\frac{\langle f(\overrightarrow{\mathrm{S_{y^*}}}), f(\overrightarrow{\mathrm{S_y}}) \rangle}{\Vert f(\overrightarrow{\mathrm{S_{y^*}}}) \Vert \Vert f(\overrightarrow{\mathrm{S_y}}) \Vert} - \frac{\langle f(\overrightarrow{\mathrm{S_{y^*}}}), f(\overleftarrow{\mathrm{S_y}}) \rangle}{\Vert f(\overrightarrow{\mathrm{S_{y^*}}}) \Vert \Vert f(\overleftarrow{\mathrm{S_y}}) \Vert}\\
\end{split}
\label{eq:reward}
\end{equation}
where $f(\overrightarrow{\mathrm{S_y}})$ is the representation of forward-ordered corresponding
gold sequence and  $f(\overleftarrow{\mathrm{S_y}})$ is the representation of reverse-ordered gold
sequence.

This reward compares the generated sequence to
both sentence orders of the gold sequence, and rewards
sequences that are more similar to the forward
order of the gold sequence. Because the cosine
similarity terms in Equation (\ref{eq:reward}) are bounded
in $[-1; 1]$, the model receives additional reward
for generating sequences that are different from
the reverse-ordered gold sequence.
\subsection{Joint Learning}
There are two decoders in our model, the first-pass decoder and the second-pass decoder, each of them can learn parameters to minimize the negative log-likelihood independently. Intuitively, these two decoders are associative, and both performance can be improved by the joint learning techniques as shown in Equation (\ref{eq:mle}). 

As aforementioned, we use a reward teacher to reward the model generating discourse coherent text. According to our architecture, there are two approaches to reward the model learning the policy. One is described in the previous section that rewards the second-pass decoder by the self-critical learning strategy. We argue that the performance can be further improved when the first-pass decoder is also rewarded by the reward teacher. Thus, the final objective of our model is to minimize:
\begin{equation}
L = L_{mle1} \cdot \lambda_1 + L_{rl1} \cdot (1-\lambda_1) + L_{mle2} \cdot \lambda_2  + L_{rl2} \cdot (1-\lambda_2)
\label{eq:final}
\end{equation}   
where $\lambda_1$ and $\lambda_2$ are two hyperparameters that balance learning the
discourse-focused policy while maintaining the accurate translation.

The losses, $L_{mle1}$, $L_{mle2}$, $L_{rl2}$ are introduced in the Equation (\ref{eq:mle1}), Equation (\ref{eq:mle2}) and Equation (\ref{eq:rl2}) respectively. The computation of $L_{rl1}$ is almost identical to the $L_{rl2}$ with slight modification by replacing the model' distribution from the first-pass decoder:
\begin{equation}
\begin{split}
L_{rl1} &= -\sum_{i}^{n}{\sum_{t}^{T_i}{R \cdot \mathrm{log}P(y_t|y_0, ..., y_{t-1}, \mathrm{H}_{enc}, \mathrm{H}_{dec1})}} \\
R &= r(\hat{y}) - r(y^*)
\end{split}
\label{eq:rl1}
\end{equation}
where $r(\hat{y})$ and $r(y^*)$ can be computed by the Equation (\ref{eq:reward}).
\section{Experiments}
We evaluate our model on the IWSLT 2015 Chinese-English translation task with TED talks as our training corpus, since it includes entire discourse text.\subsection{Data Preprocess}
\label{sec:3.1}
Considering the memory capacity, we split one talk that has more than 16 sentences into several small talks, ensuring the experiments can be successfully conducted on most GPUs. 

Specifically, we take the \textit{dev-2010} as our development set, and \textit{tst-2013$\sim$2015} as our test sets. 
Statistically, we have 14,258 talks and 231,266 sentences in the training data, 48 talks and 879 sentences in the development set, and 234 talks and 3,874 sentences in the test sets.

Following the work of \citeauthor{chaobei2017}\citeyear{chaobei2017} , we conduct byte-pair encoding \cite{DBLP:journals/corr/SennrichHB15} for both Chinese and English sentences, setting the vocabulary size to 20K and 18K respectively. 

For English tokenization, we use the script supplied by Moses Toolkit \footnote{\url{https://github.com/moses-smt/mosesdecoder/blob/master/scripts/tokenizer/tokenizer.perl}}. And we segment the Chinese sentences into words by an open source toolkit \footnote{\url{https://github.com/fxsjy/jieba}}.

\subsection{Systems}
\label{sec:3.2}
We measure the performance of our model with different system implementations. 
\begin{itemize}
\item \textit{t2t}: This is the official supplied open source toolkit for running Transformer model. Specifically, we use the v1.6.5 release \footnote{\url{https://github.com/tensorflow/tensor2tensor/releases/tag/v1.6.5}}.
\item \textit{context-encoder}: The reimplementation of the work \citeauthor{P18-1117}\shortcite{P18-1117}.\item \textit{first-pass}: This system is applied to minimize the Equation (\ref{eq:mle1}). Actually, this system is almost identical to the \textit{t2t} but with different data shuffling strategy, where the \textit{t2t} shuffles the data over the sentences randomly, while the \textit{first-pass} shuffles the data over the talks.
\item \textit{first-pass-rl}: We implement this system to minimize $L_1=L_{mle1} \cdot \lambda_1 + L_{rl1} \cdot (1-\lambda_1)$, which is the first part of the Equation (\ref{eq:final}). This system is to evaluate the policy learning strategy for the standard Transformer model.
\item \textit{two-pass}: This system is applied to minimize the Equation (\ref{eq:mle}), to measure the contribution of the two-pass decoder strategy.
\item \textit{two-pass-rl}: This is the final system, tuning to minimize the Equation (\ref{eq:final}).

\end{itemize}
We also implement the reward teacher with standard biRNNs, minimizing the Equation (\ref{eq:abs}), namely \textit{reward-teacher}.
\subsection{Training Details}
\label{sec:3.3}
Since the size of training data is relatively small in the NMT task, we use the \textit{base} version hyperparameters of the standard Transformer model, against model overfitting. For all systems, we use the Adam Optimizer \cite{kingma2014adam} with the identical settings to $t2t$, to tune the parameters. 
\begin{table*}[htb]
\begin{center}
\begin{tabular}{c|c|c|c|c}
SYSTEMS &BLEU& METEOR & BLEU$_{doc}$& METEOR$_{doc}$ \\
\hline
\textit{t2t \cite{vaswani2017attention}} & 20.10 & 35.75  & 25.60 & 34.91   \\
\textit{context-encoder} \cite{P18-1117} & 20.31 & 35.79 & 25.93 & 35.03 \\
\hline
\textit{first-pass} & 20.41 & 35.98 & 25.99& 35.23   \\
\textit{first-pass-rl} & 20.79 & 36.41 & 26.82& 36.12   \\
\textit{two-pass} & 20.92& 36.70 & 26.89& 36.29  \\
\textit{two-pass-rl} & 21.11& 36.82 & 27.50 & 36.64   \\
\hline
\textit{two-pass-bleu} & 21.08& 36.88 & 27.32 & 36.42   \\
\textit{two-pass-bleu-rl} & \textbf{21.33}& \textbf{36.94} & \textbf{27.8} & \textbf{36.89}   \\
\hline
\multicolumn{5}{c}{Pretrain with 25M bilingual training corpus from WMT2018} 	\\
\hline
\textit{t2t} & 26.57 & 42.80  & 31.65 & 39.47   \\
\textit{two-pass-bleu-rl} & \textbf{27.55}& \textbf{43.77} & \textbf{33.98} & \textbf{41.28}   \\
\end{tabular}
\end{center}
\caption{Performance of systems measured by different metrics. Due to the space limitation, we list an average score of all test datasets. To measure the discourse quality, we concatenate sentences in one talk into one long sentence, and then evaluate their BLEU and METEOR scores as BLEU$_{doc}$ and METEOR$_{doc}$.}
\label{tbl:results}
\end{table*}

One thing deserves to be noted is the value of hyperparameter \textit{batch\_size}. In general, a large value of batch size achieves better performance when training on large scale corpus (more than millions) \cite{vaswani2017attention}. However, in our preliminary experiments we found that a smaller value presented better results training on the TED talks. Thus we set the \textit{batch\_size} to 320 for \textit{t2t} system, resulting in approximately 10$\sim$20 sentences in one batch according to the different sentence length. 

For the other systems, we read one talk per one batch, producing no more than 16 sentences in one batch, which is comparable to the baseline system, \textit{t2t}.

Following the work of \citeauthor{bosselut2018discourse}\shortcite{bosselut2018discourse}, we train the \textit{reward-teacher} with the similar hyperparmaters but using different optimizing strategy. Specifically, we set both the embedding and recurrent hidden size to 100, and apply one dropout layer with keeping probability equals to 0.3 between the embedding layer and the bidirectional recurrent layers. For tuning the parameters, we use the same Adam Optimizer as the NMT systems. 

Notably, we train the \textit{reward-teacher} with the monolingual English from the TED talks as baseline system, and investigate the effect on the translation quality when the \textit{reward-teacher} trained with variant monolingual datas in later section.

The training speed of \textit{two-pass-bleu-rl} model is 8 talks per one second running on V100 with 8GPUs, and it needs about 1.5 days to converge.
\subsection{Results and Analysis}
\label{sec:3.4}
To measure the performance of our systems, we use the universal BLEU and METEOR metrics, computed by two open source toolkits \footnote{\url{https://github.com/moses-smt/mosesdecoder/blob/master/scripts/generic/multi-bleu.perl}} \footnote{\url{http://www.cs.cmu.edu/~alavie/METEOR/index.html#Download}}. In addition to measuring the quality for each individual sentence, we also concatenate sentences in one talk into one long sentence, and then represent its BLEU and METEOR scores as BLEU$_{doc}$, and METEOR$_{doc}$.

\subsubsection{Overall Results}
From the Table \ref{tbl:results}, it is interesting that our \textit{first-pass} system beats the baseline \textit{t2t} slightly. As we described in the previous section, the difference between such two systems is the diverse shuffling strategy.
Different from the \textit{t2t} system, where we shuffle the overall training data and select certain sentences into one batch randomly, while in the \textit{first-pass} system, we shuffle the talks and take sentences from one talk into one batch orderly. This finding indicates that the discourse text with ordering structure is better trained with its original order while not to be scattered, ensuring the effectiveness of our other systems since they are all trained like the \textit{first-pass} system. In addition, this training strategy can be viewed as well designed curriculum learning \cite{bengio2009curriculum} strategy, which has been proved effective for NMT task \cite{R17-1050}. 

Compared to the \textit{first-pass} system, our \textit{two-pass} system performs better results on these test datasets, which confirms that using deliberation procedure can improve the translation quality, although we implement a slightly different Deliberation Networks. 
When evaluated by the discourse metrics, we find that the Deliberation Networks is able to generate discourse coherent translation, since it can bring more average improvements compared to the sentence metrics (+0.51 vs +0.8 in term of BLEU, and +0.72 vs +1.06 in term of METEOR). 

When examining models trained using a reward
teacher, the systems (\textit{first-pass-rl} and \textit{two-pass-rl}) achieve significant improvements over these systems trained by the standard maximum likelihood estimation (\textit{first-pass} and \textit{two-pass}), by means of +0.72 BLEU$_{doc}$ and +0.62 METEOR$_{doc}$ when evaluated by the discourse metrics. Moreover, our systems can generate discourse coherent text while maintaining improving quality on each sentence, with the evidence that two systems also improve the translation quality at sentence-level when evaluated by the sentence metrics (+0.29 BLEU, +0.28 METEOR).

Another finding is that when given more context (up to 16 surrounding sentences), the \textit{two-pass} system achieves more improvements compared to the \textit{context-encoder} system which models one preceding sentence as external context. It deserves researching in the future that exploring more context to improve the translation quality.

As shown in the last row, when we take the discourse BLEU score as additional reward, our model beats the baseline system by +1.23 BLEU and +1.19 METEOR. When evaluated by discourse metrics, the improvement is more significant, by +2.2 BLEU and +1.98 METEOR, proving the effectiveness of our model. 

\subsubsection{Pretrain} 
We also investigate using a large-scale corpus, Chinese-English training corpus from the WMT2018 translation task, to pretrain the model, and then fine-tune on the TED corpus. For \textit{two-pass-bleu-rl} system, we pretrain all parameters except one special self-attention layer, which is responsible for capturing relationships between current sentence and first-pass produced coarse translations. 

From the Table  \ref{tbl:results}, it is clear that pretrained models obtain more than 6 points improvements. Also, after pretraining, our \textit{two-pass-bleu-rl} system still obtains improvements upon the baseline system, which indicates that our approach is robust and practical.
\subsubsection{Effect of Balance Factor}
We investigate the effect of two balance factors ($\lambda_1$ and $\lambda_2$ in Equation (\ref{eq:final})) on the performance of the translation quality evaluated by different type of metrics.

We first adjust the value of $\lambda_1$, ranging from 0.7 to 1.0\footnote{According to our preliminary experiments, we find that a value lower than 0.7 failed to produce reasonable translations.}, and stepping by 0.05, to see the performance change for \textit{first-pass-rl} system. Next, we then adjust the value of $\lambda_2$ with fixed value of $\lambda_1$, to optimize the performance of \textit{two-pass-rl} system.

\begin{figure}[tbp]
\centering
\includegraphics[width=3.2in]{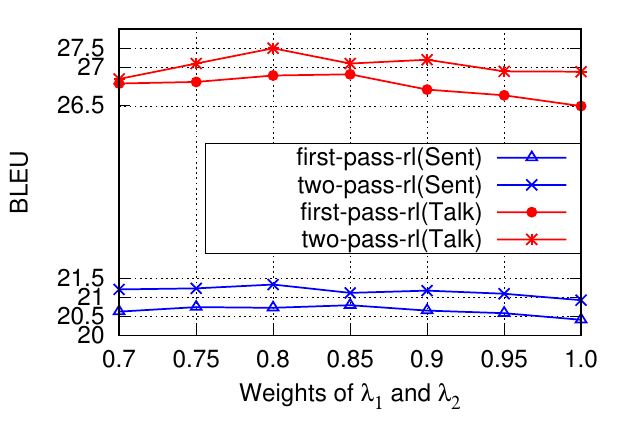}

\caption{Effect of two balance factors on the performance of the translation quality.}
\label{fig:lambda}
\end{figure}

As shown in Figure \ref{fig:lambda}, we see that setting the value of $\lambda_1$ to 0.85 and $\lambda_2$ to 0.80 produces the best performance for \textit{first-pass-rl} and \textit{two-pass-rl}.

Another finding is that the change of sentence-level BLEU score appears to be slightly consistent with the discourse BLEU$_{doc}$ score, but the latter has larger fluctuation (two top lines), indicating the reward teacher encourages generating discourse coherent translation explicitly.
\subsubsection{Effect of Reward Teacher}
Since the parameters for the \textit{reward-teacher} are fixed during the translation stage, we can explore variant reward teachers trained by different corpus, and see the effect on the translation performance. 
\begin{table}[htb]
\begin{center}
\begin{tabular}{l|c|c|c|c}
 METRICS& TED & AFP & XIN & Gigaword \\
 \hline
 BLEU & 21.11 & 20.54 & 20.92 & 20.49 \\
 BLEU$_{doc}$ & 27.50 & 26.77 & 27.32 & 26.83 \\
\end{tabular}
\end{center}
\caption{Performance of \textit{two-pass-rl} rewarded by variant reward teachers that trained by different corpus. AFP and XIN are the corresponding proportion of English Gigaword.  }
\label{tbl:reward}
\end{table}

Except the TED corpus, we train the \textit{reward-teacher} with English Gigaword Fifth Edition
\footnote{\url{https://catalog.ldc.upenn.edu/LDC2011T07}} 
that includes larger English discourse text but was written in different style. As shown in Table \ref{tbl:reward}, the teacher trained by external corpus yields bad performance, despite its large corpus size. This suggests that the reward teacher deserves to be trained with the corpus that was written in the similar style.  
\subsubsection{Coherence}
\citeauthor{lapata2005automatic}\shortcite{lapata2005automatic} proposed one approach that measures discourse coherence as sentence similarity. Specifically, the representation of each sentence is the mean of the distributed vectors of its words, and the similarity between two sentences $S_1$ and $S_2$, is determined by the cosine of their means:
\begin{equation}
sim(S_1, S_2) = \frac{\langle f(S_1), f(S_2) \rangle}{\Vert f(S_1) \Vert \Vert f(S_2) \Vert}
\label{eq:coherence}
\end{equation}
where $f(S_i) = \sum_{w \in S_i} \vec{w}$, and $\vec{w}$ is the vector for word $w$.

We use Word2Vec 
\footnote{\url{http://word2vec.googlecode.com/svn/trunk/}} 
to learn the distributed vectors of words by training on the aforementioned English Gigaword Fifth Edition. And we set the dimensionality of word embeddings to 100.
\begin{table}[htb]
\begin{center}
\begin{tabular}{c|c|c|c}
SYSTEMS & \textit{tst-2013} & \textit{tst-2014} & \textit{tst-2015} \\
\hline
\textit{t2t} &  0.5991& 0.5838& 0.5939\\
\textit{first-pass} &0.5999& 0.5845& 0.5943\\
\textit{two-pass} &0.6011& 0.5880& 0.5962\\
\textit{first-pass-rl} &0.6008& 0.5861& 0.5952\\
\textit{two-pass-rl} &0.6032& 0.5913& 0.6008\\
\textit{two-pass-bleu-rl} &0.6041& 0.5938& 0.6014\\
\textit{human translation} & 0.6066 &0.5910 &  0.6013\\
\end{tabular}
\end{center}
\caption{We measure the discourse coherence as sentence similarity.}
\label{tbl:coherence}
\end{table}

Table \ref{tbl:coherence} shows the cosine similarity of adjacent sentences on all test datasets. It reveals that systems encouraged by discourse reward produce better coherence in document translation than contrastive systems in term of cosine similarity. 
\subsubsection{Conjunctions}
We count the top five frequent conjunctions in the translations produced by \textit{t2t} and \textit{two-pass-rl}, to see the concrete transformation of sentences that encouraged to generate coherent translations.
\begin{table}[htb]
\begin{center}
\begin{tabular}{l|l}
\textit{t2t} & And (519) But (186) In (114) So (174) What (55) \\
\hline
\textit{sys$^*$} & And (540) But (183) In (129) So (178) What (73)
\end{tabular}
\end{center}
\caption{The statistics of top five frequent conjunctions in two systems (\textit{sys$^*$} is \textit{two-pass-bleu-rl}). Numbers in bracket is the  occurrences of this word.  }
\label{tbl:conj}
\end{table}

 As shown in Table \ref{tbl:conj}, sentences in \textit{two-pass-bleu-rl} tend to using more diverse conjunctions to build the connections towards preceding sentence, which proves that our model can generate more discourse coherent translation. 

\section{Related Work}
\citeauthor{gong2011cache}\shortcite{gong2011cache} proposed a memory based approach to capture contextual information to facilitate the statistical translation model generating discourse coherent translations, and the literatures \cite{kuang2017cache,tu2018learning,P18-1118} extended similar memory based approach to the NMT framework.

\citeauthor{wang2017exploiting}\shortcite{wang2017exploiting} presented a novel document RNN to learn the representation of the entire text, and treated the external context as the auxiliary context which will be retrieved by the hidden state in the decoder.
 
 \citeauthor{tiedemann2017neural}\shortcite{tiedemann2017neural} and \citeauthor{P18-1117}\shortcite{P18-1117} proposed to encode global context through extending the current sentence with one preceding adjacent sentence. Notably, the former was conducted on the recurrent based models while the latter was implemented on the Transformer model. 
\section{Conclusion and Future Work}
In this paper, we propose two novel techniques, Deliberation Networks and reward teacher to generate discourse coherent translation. Practical experiments confirm, through modeling external discourse context from the potential translations of the other sentences in the same text, our model can improve the translation quality both on the sentence-level (+1.23 BLEU) and discourse-level (+2.2 BLEU) metrics. Moreover, when the model learns the policy that rewarded by a reward teacher, it can generate more fluent and coherent discourse translations.
In the future, we will continue research on using more bilingual training data that has no explicit discourse boundaries, and verify our model on multi-lingual  translation tasks.

\bibliographystyle{aaai19} 
\bibliography{aaai19-xionghao}
\clearpage
\end{CJK}
\end{document}